\documentclass{article}
\usepackage{spconf,amsmath,graphicx}
\usepackage{booktabs}
\usepackage[flushleft]{threeparttable}
\usepackage[utf8]{inputenc}

\usepackage[capitalize]{cleveref}
\Crefname{section}{Sec.}{Secs.}
\Crefname{section}{Section}{Sections}
\Crefname{table}{Table}{Tables}
\Crefname{table}{Tab.}{Tabs.}

\title{Recovering from Privacy-Preserving Masking with Large Language Models}
\name{
    Arpita Vats$^{1}$, Zhe Liu$^{2}$, Peng Su$^{2}$, Debjyoti Paul$^{2}$, Yingyi Ma$^{2}$,
    \emph{Yutong Pang}$^{2}$, \emph{Zeeshan Ahmed}$^{2}$, \emph{Ozlem Kalinli}$^{2}$
    \thanks{* Work done during an internship at Meta.}
}
\address{
    $^{1}$Santa Clara University, Santa Clara, CA,\\ $^{2}$Meta, Menlo Park, CA
}

\begin{document}
\ninept
\maketitle
\begin{abstract}
Model adaptation is crucial to handle the discrepancy between proxy training data and actual users' data received. To effectively perform adaptation, textual data of users is typically stored on servers or their local devices, where downstream natural language processing (NLP) models can be directly trained using such in-domain data. However, this might raise privacy and security concerns due to the extra risks of exposing user information to adversaries. Replacing identifying information in textual data with a generic marker has been recently explored. In this work, we leverage large language models (LLMs) to suggest substitutes of masked tokens and have their effectiveness evaluated on downstream language modeling tasks. Specifically, we propose multiple pre-trained and fine-tuned LLM-based approaches and perform empirical studies on various datasets for the comparison of these methods. Experimental results show that models trained on the obfuscation corpora are able to achieve comparable performance with the ones trained on the original data without privacy-preserving token masking.

%In the digital era, ensuring privacy is of utmost importance and it will yield high risk of privacy disclosure when storing personal data on servers. More and more applications start using \texttt{<mask>} token to hide the private information from users. However, it will pose a big challenge on the downstream machine learning (ML) applications based on the data from users, due to the incomplete context of the training data. In this work, we explore the methods of compensating the unavailability of the privacy information on the machine learning model training. More specifically, we employ the large language models (LLMs) for imputing the unavailable words (privacy information) in training data. Through comprehensive experiments, we verify the effectiveness of our approach on improving the model performance. Additionally, we also assess the potential of our proposed method in speech domain, by applying a trained neural language model in the shallow fusion of the Automated Speech Recognition (ASR) model.

\end{abstract}
\begin{keywords}
Privacy-preserving machine learning, language modeling, large language models, automatic speech recognition
\end{keywords}
\section{Introduction}
\label{sec:intro}
A common issue arising after deploying a machine learning model on central servers or user devices is the discrepancy between training data and actual user data received. Specifically, in the applications of natural language processing (NLP), semantic characteristics and topics of real users' textual data could be very different from those of server-side proxy corpora, in which scenarios model {adaptation} is indispensable \cite{li2020empirical, LiuLiBak2021}.

To effectively perform model adaptation, textual data of users is typically stored on servers or their devices, where any downstream NLP models will be trained using such in-domain data. However, users' personal data might contain sensitive user information, such as people's names, addresses, and credit card numbers. Therefore, this conventional practice of users' data storage might raise privacy and security concerns due to the risks of exposing user information to adversaries. In addition, recent research has shown that sensitive information in training datasets can be detected and then extracted in unexpected ways \cite{fredrikson2015model, song2019auditing, carlini2019secret, carlini2021extracting, huang2022detecting}. Particularly, language models (LMs) are prone to {unintentionally memorize} rare or unique sequences of data, and when being prompted appropriately, they will be able to emit the memorized text verbatim \cite{carlini2022quantifying}. Thus, having NLP models directly trained on private user data might have extra risks of exposing sensitive information.

To overcome these challenges, replacing identifying information in textual data with a generic marker has been explored ~\cite{martinez2012privacy,sousa2023keep,preiss2023automatic}. To be more specific, tokens considered as sensitive or private are masked out using some special symbol, such as ``[MASK]''. In the example where the raw textual sequence is ``Tom lives in Chicago'', one might mark the words of ``Tom'' and ``Chicago'' as personal and thus replace them with the mask symbol. The resulting sequence is ``[MASK] lives in [MASK]'', which will be stored into servers or local devices for model adaptation purposes later on.

While this strategy is capable to provide privacy protections on user data, it also introduces significant complexities to the training of any NLP models for downstream adaptation tasks. The existence of markers might break the semantic structures, disrupt the coherence of languages, or fail to preserve the meaning of the original textual sequences. As a result, models directly trained on the masked corpus could yield much worse performance compared with the ones trained on the raw corpus without privacy-preserving token masking. Therefore, it calls for advanced approaches on effectively substituting the masked tokens in the corpus and bridge the accuracy gaps in NLP models for adaptation tasks.

In this work, we propose to use large language models (LLMs) to provide appropriate candidate tokens to fill in the generic markers in any masked corpus. Note that predicting the masked tokens based on the surrounding context can be considered as a task of masked LM (MLM), thus bi-directional Transformer \cite{vaswani2017attention} based pre-trained LLMs, such as BERT \cite{devlin2018bert} and RoBERTa \cite{liu2019roberta}, would be suitable for this endeavor. Upon observing the remarkable capabilities demonstrated by decoder-only LLMs, models such as ChatGPT \cite{chatgpt} and LLaMA2 \cite{touvron2023llama} can also be utilized here for providing substitutes of masked tokens. Our goal is not to restore any markers to the original tokens without masking, instead, we aim to replace any masked token with some substitute of the same type. More specifically, the efficiency of any recovering method from privacy-preserving masking shall be evaluated on the downstream adaptation tasks, through the NLP models trained on the obfuscation corpus. In this paper, we use language modeling and LM-fused automatic speech recognition (ASR) \cite{mikolov2010recurrent, chen2015improving, liu2014efficient, kannan2018analysis, irie2019language} as the downstream tasks. 

We make the following contributions:
\begin{itemize}
    \item To the best of our knowledge, our work is the first to leverage LLMs to suggest substitutes of masked tokens and have their effectiveness evaluated on downstream LM and ASR tasks;
    \item We propose multiple pre-trained and fine-tuned LLM-based methods and conduct empirical experiments on various NLP datasets for the comparison of adapted models accordingly. The results of our experiments indicate that models trained on the obfuscation corpora have comparable performance with the ones trained on the original data without privacy-preserving token masking;
    \item We also present three token masking techniques and measure the performance of our proposed methods on each of them in downstream tasks as well.
\end{itemize}
The rest of the paper is organized as follows. We review related works in Section~\ref{sec:related}. Section \ref{sec:method} describes the details of our proposed framework on privacy-preserving token
masking and the substitutes of masked tokens using LLMs. Next, Section \ref{sec:expt} shows the experiments and results for downstream tasks of LM and ASR. Finally, We conclude in Section \ref{sec:conclude}.

\section{Related Works}
\label{sec:related}
Privacy protection has been becoming crucial in NLP research \cite{sousa2023keep}. One important direction in this area is through anonymization, which involves the removal of identifying information from textual corpus \cite{martinez2012privacy, lison2021anonymisation, hartman2020customization}. More recently, obfuscation, replacing any sensitive information with a different substitute of the same type has been investigated. In particular, a survey of profanity obfuscation in NLP is conducted in \cite{nozza2022state}. Authors in \cite{Zhifeng} employs a neural model that aims to preserve the syntactic relationships of the original sentence so that the obfuscated sentence can be parsed instead of the original one; it outperforms random substitution baselines across syntactic parsers. The work of \cite{preiss2023automatic} studies named entity obfuscation in speech, which focuses on identifying, replacing, and inserting replacement named entities synthesized using voice cloning into original audio. The paper of \cite{blau2023using} improves the speech recognition of personal identifiers by including fake textual substitutes in the training data of ASR. None of these existing works explore the use and comparison of different LLMs for suggesting token substitutes in obfuscation.

\section{Methodology}
\label{sec:method}
We describe our proposed approaches on privacy-preserving token masking and the substitutes of masked tokens using LLMs. Specifically, we introduce several token masking techniques in Section~\ref{method:masking}; LLM-based methods on replacing the masked tokens are presented in Section~\ref{method:recovery}; Section~\ref{method:downstream} discusses the use of obfuscation corpus for performing language modeling task.

\begin{figure}[htb]
\begin{minipage}[b]{1.0\linewidth}
  \centering
  \centerline{\includegraphics[width=8.5cm]{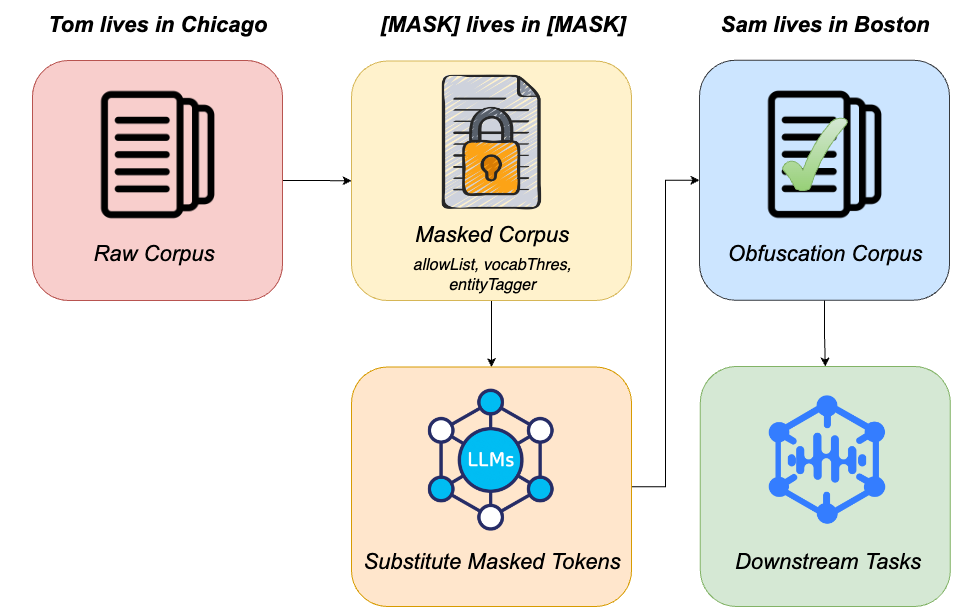}}
  % \vspace{2.0cm}
  \caption{\label{fig_pipeline} The framework of token masking and obfuscation using LLMs.}
\end{minipage}
\end{figure}

The overall framework is depicted in Figure~\ref{fig_pipeline}.

% In this section, we first explore several privacy masking methods to generate privacy-masked datasets, since there is no such public dataset available. Then we introduce the proposed method of applying large language model on recovering masked words. At last, the recovered dataset will be used to train a downstream machine learning model, and the evaluation results will be seen as the metric to measure the quality of the recovered data. The whole pipeline is shown in Fig.~\ref{fig_pipeline}. In this work, we will train a language model (LM) for the downstream application, and the evaluation results on this LM will be seen as the measurement metrics of our proposed methods. 
\subsection{Token Masking Techniques}
\label{method:masking}
%Oftentimes it is expensive to manually label the sensitive tokens in a corpus of user data. To overcome this problem, we employ various masking techniques to mask the information for us automatically. More specifically, our exploration encompasses three different methods for masking information:
Masking sensitive tokens from users' data helps reduce the privacy risks and prevent any personal information being leaked or extracted from adversaries. Such token masking task shall be performed without human-in-the-loop since practitioners are not allowed to have the access to annotate or label private data of users.

To automatically conceal sensitive information in some private corpus, we propose the following token masking techniques:
\begin{itemize}
    \item $allowList$: This is a pre-defined list of tokens that are considered non-sensitive and safe to keep. Typically, such list is handcrafted by linguistic specialists. Then during the process of masking, any token not present in this allow list will be masked out;  %This compilation comprises 5,000 commonly encountered words, thoughtfully curated by the linguist team at Meta. It encompasses words commonly used in everyday conversations. We employ the $mask$ token to conceal any term that does not belong to the $allowList$. In essence, any word not present in the $allowList$ is masked.
    \item $vocabThres$: This involves the selection of $N$ most frequent tokens from a vocabulary as the list of non-sensitive tokens. That is, any token with its frequency less than some threshold will be masked out. Here, the vocabulary set can be built from some generic large corpora; %Subsequently, a similar technique with $allowList$ is applied: words present in the $vocabList$ are not masked.
    \item $entityTagger$: In this approach, named entity recognition (NER) models are utilized to identify potential entities in any private corpus, which will be treated as personal tokens and masked out. These entities include but are not limit to individuals' names, locations, and organizations.
\end{itemize}
Throughout these masking techniques, we will more likely mask the non-common tokens in any corpus, assuming privacy information is more related to rare or unique tokens. After applying the masking, we obtain a masked corpus where the masked tokens were replaced with the symbol of ``[MASK]''.

\subsection{Recovery Methods from Masking}
\label{method:recovery}
Token masking provides privacy protections, however, the resulting masked corpus might not be suitable to be directly used for training NLP models for downstream tasks.

Given any masked corpus, we propose to use LLMs to fill in each mask symbol with appropriate token that matches the semantic contexts. It is important to note that we are not aiming to predict exactly the same token with the original one in the raw corpus. We expect to substitute it with some token that makes the whole sentence linguistically correct and complete.

The following illustrates different strategies on leveraging LLMs for substituting masked tokens: %To fully harness the power the LLMs, we employ different strategies to utilize the outputs from LLMs for our task. More details are as follows and also demonstrated in Fig.~\ref{fig_proposed}.

\begin{itemize}
\item \texttt{Top-1}: In this method, we directly use the 1-best predicted token from an LLM to replace the masked token. Here, token filling is considered as a masked LM task. If there are multiple markers in the sentence, they are replaced in a sequential order from the left to the right, one at a time;

\item \texttt{Top-K}: This approach extends the token filling candidates from the 1-best to the $K$-best from the predictions of an LLM. Specifically, we randomly choose a token from the top-$K$ predictions. %and if the chosen token is contained in the set of all masked tokens, we repeat the sampling process until the condition is met. 
Then this selected token is used to fill in the marker in the sentence. For substituting any masked tokens from $allowList$ or $vocabThres$ based masking techniques, we prefer the predicted tokens not being included in the corresponding token list, thus we repeat the random sampling process until this condition is met or there is no available candidates of predicted tokens among the top-$K$;

\item \texttt{Fine-Tuning(FT)}: In the previous two approaches, we utilize the token predictions from a pre-trained LLM. Fine-tuning a pre-trained LLM using in-domain corpus helps the model gain domain-specific knowledge, and hence enhance the performance in the masked token prediction. To accomplish this, samples without any masked tokens can be used for fine-tuning. However, in many scenarios, it is possible that majority of samples contain at least one mask symbol so that fine-tuning is less effective especially when the size of corpus is small. Alternatively, the top-1 or top-$K$ predictions from the same pre-trained LLM can be firstly used to substitute the masked tokens in any samples, and then the entire obfuscation corpus can be used for fine-tuning the LLM. Once we have a fine-tuned LLM, either \texttt{Top-1} or \texttt{Top-K} can be applied for the substitution of masked tokens. Note that the process above can be utilized for multiple times.
\end{itemize}
After applying any of these methods, we obtain an obfuscation corpus that does not contain any masks.

\iffalse
\begin{figure}[htb]
\begin{minipage}[b]{1.0\linewidth}
  \centering
  \centerline{\includegraphics[width=7.5cm]{image/proposed_methods.drawio.png}}
%  \vspace{2.0cm}
  \caption{ \label{fig_proposed} Proposed method pipeline.}
\end{minipage}
\end{figure}
\fi
\subsection{Performing Downstream Tasks}
\label{method:downstream}
Once we have substituted masked tokens, the resulting corpus can be used for training machine learning models for any downstream tasks. Notice that the effectiveness of any token filling approach should be measured by the performance of these machine learning models on these downstream tasks.

In this work, we consider the language modeling adaptation task where a generic pre-trained LM is fine-tuned on the obfuscation corpus. This adapted LM will be evaluated on a (unmasked) test set which has the same domain with the raw corpus. The performance of LM is measured in term of perplexity.

When integrating an adapted LM with an ASR model via shallow fusion, word error rate (WER) can also be evaluated on a test set of utterances. %we need to evaluate the quality of the data. In this work, we use the performance of the downstream machine learning model as the metric of the recovered data quality from LLM. Specifically, we train a small language model on the recovered dataset and evaluate it on the respective test sets.

% As for the evaluation results, we utilize two different metrics: a) the perplexity of the trained language model will reflect the model quality in general, so we use it as our principle metric; b) we also apply the language model in shallow fusion of the ASR model to measure how much the language model could boost the ASR model performance. Thus the Word Error Rate (WER) for measuring the ASR model performance will be our secondary metric.  

\section{Experiments}
\label{sec:expt}

\subsection{Datasets}
To compare the performance of multiple baselines and our proposed approaches on the downstream language modeling task, we explore three datasets in the experiments: Fisher~\cite{fisher}, Pushshift.io Reddit\footnote{Pushshift.io Reddit dataset is a previously existing dataset extracted and obtained by a third party that contains preprocessed comments posted on the social network Reddit and hosted by pushshift.io. We will refer this dataset as ``Reddit'' in the rest of the paper.}~\cite{reddit}, and Wall Street Journal (WSJ)~\cite{WSJ}. The  statistics of these datasets are summarized in \cref{tbl:dataset}. The test set of WSJ data also consists of voice utterances and is thus used for evaluating the ASR models with fused LMs.

\begin{table}[ht!]
  \caption{Data information.}
  \label{tbl:dataset}
  \centering
  \resizebox{0.7\columnwidth}{!}{%
  \begin{threeparttable}
  \begin{tabular}{l|r|r}
    \toprule
    & \emph{Train Set (\#sent)} & \emph{Test Set (\#sent)} \\
    \midrule
    Fisher & 1,158,496 & 50,000\\
    Reddit & 763,683 & 49,570 \\
    WSJ & 6,000 & 800 \\
    \bottomrule
  \end{tabular}
  \end{threeparttable}
  }
\end{table}

\subsection{Setups}
\subsubsection{Downstream Tasks}
The downstream LM is a Transformer with 6 layers, 12 attention heads, and 768 hidden units. The set of word vocabulary is around 85K. The LM is pre-trained on WikiText-103 corpus \cite{merity2016pointer}.

For each of the masking techniques considered in this study, LMs are fine-tuned on the obfuscation train sets of Fisher, Reddit, and WSJ data. Their perplexities are evaluated on the corresponding test sets.

On the WSJ test set, we also evaluate the ASR performance. The ASR model is an RNN-T model with the Emformer encoder \cite{emformer2021streaming}, LSTM predictor, and a joiner. It has around 80 million parameters and is trained from scratch using the train split of LibriSpeech ASR corpus \cite{panayotov2015librispeech}.

\subsubsection{Masking Techniques}
In our experiments, $allowList$ contains a set of 5K curated common words, and $vocabThres$ consists of 10K most frequent words among the same 85K word vocabulary mentioned above. For the $entityTagger$ masking technique, we utilize the BERT-NER model \cite{devlin2018bert, kim2003introduction} for tagging named entities in the train sets. 

For each of these masking techniques, \cref{tbl:mask} shows the percentage of masked tokens per dataset. We can see that $allowList$ masks many more tokens than the other two techniques.
\begin{table}[ht!]
  \caption{Percentages of masked tokens.}
  \label{tbl:mask}
  \centering
  \resizebox{0.75\columnwidth}{!}{%
  \begin{threeparttable}
  \begin{tabular}{l|r|r|r}
    \toprule
     & \emph{allowList} & \emph{vocabThres} & \emph{entityTagger} \\
    \midrule
    Fisher & 12.5\% & 1.3\% & 1.7\% \\
    Reddit & 22.7\% & 11.9\% & 4.2\% \\
    WSJ & 30.4\% & 11.2\% & 9.1\% \\
    \bottomrule
  \end{tabular}
  \end{threeparttable}
  }
\end{table}

\subsubsection{Baselines}
We consider the following methods as the baselines:
 \begin{itemize}
   \item \texttt{Oracle:} an LM is trained on the ground-truth sentences without any masking, which provides the upper bound for the model performance on each dataset;
    \item \texttt{Baseline0:} an LM is directly trained on the masked corpus, where the mask symbol ``[MASK]'' is treated as a special token during model training;
    \item \texttt{Baseline1:} zero weight is assigned to any mask symbol ``[MASK]'' in the LM loss function during model training.
 \end{itemize}
Note that for each of these methods, the LM is still pre-trained on the WikiText-103 corpus. 

\subsubsection{LLM-Based Methods}
In our experiments, we consider the following LLMs for substituting masked tokens in any training sequences: \texttt{BERT} (base, uncased), \texttt{RoBERTa} (base), and \texttt{LLaMA2} (7B model parameters).

For the fine-tuning of \texttt{BERT} and \texttt{RoBERTa}, we use MLM as the training task. During the inference time of using pre-trained or fine-tuned \texttt{BERT} and \texttt{RoBERTa} to substitute masked tokens, any consecutive markers of ``[MASK]'' are merged into one marker. We set $K=10$ in the \texttt{Top-K} method.

For \texttt{LLaMA2}, we adopt a different approach for the fine-tuning process since it is an auto-regressive model. Specifically, for each training sample, we generate prompts by combining some instruction, input, and output text: instruction contains the text of ``Predict the [MASK] tokens in the given sentence''; input is the same training sample but having a few tokens randomly replaced with the symbol of ``[MASK]''; and output is the original training sample (without masking). We leverage the low-rank adaptation (LoRA) method \cite{hu2021lora} for fine-tuning \texttt{LLaMA2} on the set of prompts. During the inference time, the instruction and input are provided to the fine-tuned model, which allows the model for continued text generation.

\subsection{Results}
\label{sec: Results}

Table~\ref{tbl:fisher} shows the perplexity results of the baselines and proposed methods on Fisher dataset. We have the following observations:
\begin{itemize}
\item All proposed methods give lower perplexity results than the two baseline methods;
\item In all scenarios, \texttt{Top-K} outperforms \texttt{Top-1} based methods; fine-tuned \texttt{BERT} and \texttt{RoBERTa} obtain better results than the ones without fine-tuning;
\item Since more tokens are masked out with $allowList$, the  gap between \texttt{Oracle} and any other method is much larger than that of $vocabThres$ or $entityTagger$ masking technique;
\item \texttt{RoBERTa} yields the best perplexity performance across all the masking techniques. In particular, for $vocabThres$ and $entityTagger$, perplexity results from fine-tuned \texttt{RoBERTa} are very close to those of \texttt{Oracle}, which indicates that most of the missing information can be recovered in the obfuscation dataset;
\item \texttt{LLaMA2(Top-1,FT)} is a competitive method but is not as good as fine-tuned \texttt{BERT} or \texttt{RoBERTa} for this task.
\end{itemize}

\begin{table}[ht!]
  \vspace{-0.4cm}
  \caption{Perplexity results on Fisher dataset.}
  \label{tbl:fisher}
  \centering
  \resizebox{0.9\columnwidth}{!}{%
  \begin{threeparttable}
  \begin{tabular}{l|r|r|r}
    \toprule
     & \emph{allowList} & \emph{vocabThres} & \emph{entityTagger} \\
    \midrule
\texttt{Oracle} & 37.3 & 37.3 & 37.3  \\
\midrule
\texttt{Baseline0} & 120.1 & 42.3 & {41.7} \\
\texttt{Baseline1} &109.4 & {41.6} & {41.6} \\
\midrule
\texttt{BERT(Top-1)} & 93.0 & {41.3}& 41.5\\
\texttt{RoBERTa(Top-1)} & 71.6 & 40.5& {39.5} \\
\midrule
\texttt{BERT(Top-K)} &75.2 &40.8&{40.5}\\
\texttt{RoBERTa(Top-K)}& 70.2&\textbf{38.9}&{38.7}\\
\midrule
\texttt{BERT(Top-K,FT)} &73.6 &{39.8}&{39.7}\\
\texttt{RoBERTa(Top-K,FT)}& \textbf{65.3}&\textbf{38.9}&\textbf{38.5}\\
\midrule
\texttt{LLaMA2(Top-1,FT)} &89.3 &40.8& {40.7}\\
    \bottomrule
  \end{tabular}
  \end{threeparttable}
  }
\end{table}

\cref{tbl:reddit} shows the experimental results on Reddit dataset. The observations are similar to the ones in Fisher dataset. In particular, \texttt{RoBERTa(Top-K,FT)} again achieves the best perplexity results across all the masking techniques. 

\begin{table}[ht!]
  \vspace{-0.4cm}
  \caption{Perplexity results on Reddit dataset.}
  \label{tbl:reddit}
  \centering
  \resizebox{0.9\columnwidth}{!}{%
  \begin{threeparttable}
  \begin{tabular}{l|r|r|r}
    \toprule
     & \emph{allowList} & \emph{vocabThres} & \emph{entityTagger} \\
    \midrule
\texttt{Oracle}& 76.0 &76.0 &76.0  \\
\midrule
\texttt{Baseline0} & 339.6 & 168.2 &{82.3} \\
\texttt{Baseline1} &221.9   & 134.9 & {79.8} \\
\midrule
\texttt{BERT(Top-1)} & 196.2 & 121.2& {78.9}\\
\texttt{RoBERTa(Top-1)}& 117.3 & 94.2& {78.4} \\
\midrule
\texttt{BERT(Top-K)}&127.4 &106.3&{78.7}\\
\texttt{RoBERTa(Top-K)}&123.4&92.6&{77.4}\\
\midrule
\texttt{BERT(Top-K,FT)} &117.4&102.5& {77.6}\\
\texttt{RoBERTa(Top-K,FT)}& \textbf{98.5}&\textbf{82.1}&\textbf{76.8}\\
\midrule
\texttt{LLaMA2(Top-1,FT)}&123.3 &107.7&{78.7}\\
    \bottomrule
  \end{tabular}
  \end{threeparttable}
  }
\end{table}

\cref{tbl:wsj} and \cref{tbl:wsj_wer} show the perplexity and WER results on WSJ dataset, respectively. We have the following findings:
\begin{itemize}
\item The use of fused LM for conducting domain adaptation in ASR models is effective: comparing the WERs between ASR models with the pre-trained LM and the \texttt{Oracle} LM, there is a more than 15\% WER improvement achieved by the latter;
\item The best WERs obtained by proposed methods have relatively small gaps compared with those of the \texttt{Oracle} LM. For  $vocabThres$ and $entityTagger$ masking techniques, the WERs from \texttt{Oracle} are lifted by only 1\% (10.7 versus 10.6) and 5\% (11.1 versus 10.6), respectively. That is, the proposed methods are able to achieve significant improvements over the pre-trained LM (without adaptation), while they also provide better privacy protection than the \texttt{Oracle} LM.
\end{itemize}

\begin{table}[ht!]
  \vspace{-0.4cm}
  \caption{Perplexity results on WSJ dataset.}
  \label{tbl:wsj}
  \centering
  \resizebox{0.9\columnwidth}{!}{%
  \begin{threeparttable}
  \begin{tabular}{l|r|r|r}
    \toprule
     & \emph{allowList} & \emph{vocabThres} & \emph{entityTagger} \\
    \midrule
\texttt{Oracle}& 86.5 &86.5 &86.5  \\
\midrule
\texttt{Baseline0} & 309.0 & {144.3} & 204.0 \\
\texttt{Baseline1} &210.0   & {122.9} & 198.2 \\
\midrule
\texttt{BERT(Top-1)}& 205.9 & {119.4}& 149.3\\
\texttt{RoBERTa(Top-1)} & 181.1 &{102.5}&118.2 \\
\midrule
\texttt{BERT(Top-K)}&174.1 &103.3&108.3\\
\texttt{RoBERTa(Top-K)}&\textbf{114.5}&\textbf{93.4}&\textbf{98.7}\\
\midrule
\texttt{BERT(Top-K,FT)} &186.7&113.4& 162.3\\
\texttt{RoBERTa(Top-K,FT)}& 120.7&110.4&157.8\\
\midrule
\texttt{LLaMA2(Top-1,FT)}&135.6 &106.8&145.6\\
    \bottomrule
  \end{tabular}
  \end{threeparttable}
  }
\end{table}

\begin{table}[ht!]
  \vspace{-0.4cm}
  \caption{WER results on WSJ dataset.}
  \label{tbl:wsj_wer}
  \centering
  \resizebox{0.9\columnwidth}{!}{%
  \begin{threeparttable}
  \begin{tabular}{l|r|r|r}
    \toprule
     & \emph{allowList} & \emph{vocabThres} & \emph{entityTagger} \\
    \midrule
\texttt{ASR-without-LM} & 14.4 & 14.4 & 14.4\\
\texttt{Pre-Trained-LM} & 12.6 & 12.6 & 12.6\\
\midrule
\texttt{Oracle}& 10.6&10.6&10.6  \\
\midrule
\texttt{Baseline0} & 13.0 & 12.6 & {11.3} \\
\texttt{Baseline1}&12.5  &{11.2}  &{11.2}  \\
\midrule
\texttt{BERT(Top-1)} &12.4  & {11.1}&11.2 \\
\texttt{RoBERTa(Top-1)} &12.4  &{10.9} &\textbf{11.1} \\
% \textbf{Top 1(Llama2-chat) } &  &  & \\
\midrule
\texttt{BERT(Top-K)}&12.1 &{11.1}&11.4\\
\texttt{RoBERTa(Top-K)}&11.9&10.9&\textbf{11.1}\\
\midrule
\texttt{BERT(Top-K,FT)} &12.7&{11.5}&11.7\\
\texttt{RoBERTa(Top-K,FT)}&\textbf{11.8}&11.4&\textbf{11.1} \\
\midrule
\texttt{LLaMA2(Top-1,FT)}&12.0&\textbf{10.7}&{11.2}\\
    \bottomrule
  \end{tabular}
  \end{threeparttable}
  }
\end{table}

\section{Conclusion}
\label{sec:conclude}
In this paper, we propose multiple pre-trained and fine-tuned LLM-based methods to recover from privacy-preserving token masking on textual corpus and perform empirical studies on various datasets for the comparison of these approaches. Our experimental results demonstrate that LMs trained on the obfuscation corpora can obtain comparable accuracy with the ones trained on the raw data without privacy-preserving token masking.

Future research might include fine-tuning LLMs with the object function designed to be more directly related to the downstream NLP tasks. Also, we would consider a combination of these three masking techniques and adopt class-specific markers such as ``[PERSON]'', ``[NUMBER]'', etc. 

% Working on it
%In this work, we introduce a new research topic about privacy-preserving machine learning, and design comprehensive experiments to explore the effect of privacy information on the downstream machine learning model training. We also propose the methods of utilizing LLMs for imputing the masked token (privacy information) for improving the downstream model performance. The experiment results on three different datasets prove the effectiveness of our methods. In the future, we plan to keep exploring different LLMs for the applications in the privacy-preserving machine learning area. Additionally, we aim to further investigate different methods for effective privacy masking.

% \section{REFERENCES}
% \label{sec:refs}

% \bibliographystyle{IEEEbib}
% \bibliography{main}

\bibliographystyle{IEEEbib}
\footnotesize
\bibliography{main}

\end{document}